%% file: paper.tex
\pdfoutput=1

\documentclass[11pt]{article}
\usepackage{ACL2023}

\usepackage{times}
\usepackage{latexsym}
\usepackage[T1]{fontenc}
\usepackage[utf8]{inputenc}

\usepackage{microtype}
\usepackage{inconsolata}

\usepackage{cite}
\usepackage[pdftex]{graphicx}
\usepackage[cmex10]{amsmath}
\usepackage{algorithm}
\usepackage{algorithmicx}
\usepackage{algpseudocode}
\usepackage{mdwmath}
\usepackage{mdwtab}
\usepackage{amssymb}
\usepackage{amsmath}
\usepackage{wasysym}
\usepackage{subfigure}
\usepackage{tabularx}
\usepackage{mdwlist}
\usepackage{url}
\usepackage{multirow}

\DeclareMathAlphabet{\mathitbf}{OML}{cmm}{b}{it}

\newcommand{\ie}{\emph{i.e.}}

\renewcommand{\Huge}{\fontsize{23.5pt}{27pt}\selectfont}

\newcommand*{\horzbar}{\rule[.5ex]{2.5ex}{0.5pt}}

\title{Is Cross-modal Information Retrieval\\Possible without Training?}

\author{Hyunjin Choi\,~~~Hyunjae Lee\,~~~Seongho Joe\,~~~Youngjune L. Gwon\\
Samsung SDS}

\begin{document}
\maketitle

\begin{abstract}
Encoded representations from a pretrained deep learning model (e.g., BERT text embeddings, penultimate CNN layer activations of an image) convey a rich set of features beneficial for information retrieval. Embeddings for a particular modality of data occupy a high-dimensional space of its own, but it can be semantically aligned to another by a simple mapping without training a deep neural net. In this paper, we take a simple mapping computed from the least squares and singular value decomposition (SVD) for a solution to the Procrustes problem to serve a means to cross-modal information retrieval. That is, given information in one modality such as text, the mapping helps us locate a semantically equivalent data item in another modality such as image. Using off-the-shelf pretrained deep learning models, we have experimented the aforementioned simple cross-modal mappings in tasks of text-to-image and image-to-text retrieval. Despite simplicity, our mappings perform reasonably well reaching the highest accuracy of 77\% on recall@10, which is comparable to those requiring costly neural net training and fine-tuning. We have improved the simple mappings by contrastive learning on the pretrained models. Contrastive learning can be thought as properly biasing the pretrained encoders to enhance the cross-modal mapping quality. We have further improved the performance by multilayer perceptron with gating (gMLP), a simple neural architecture. 
\end{abstract}

\input{intro}
\input{related}
\input{approach}
\input{exp}
\input{conc}

\bibliography{paper}
\bibliographystyle{acl_natbib}

\end{document}

%% file: intro.tex
\section{Introduction}
Cross-modal information retrieval takes in one modality (or type) of data as a query to retrieve semantically related data of another type. There is a fundamental challenge in measuring the similarity between the query and the outcome having different modalities. Research in cross-modal retrieval has naturally focused on learning or training a joint subspace where different modalities of data can be compared directly. 

Recently, pretraining deep learning models with large-scale data has proved effective for creating applications in computer vision and natural language processing (NLP). Available publicly, pretrained models are a powerful encoder of characteristic features for data onto an embedding space. Pretrained models are valid in a unimodal scenario, and it is difficult to purpose them for cross-modal (or multimodal) usage. Joint training of different data modalities in a large scale would be extremely difficult and costly (or it may not be feasible). 

In this paper, we compute a simple mapping instead for cross-modal translation via the least squares and singular value decomposition (SVD). Embedding representations from a pretrained unimodal encoder are aligned \emph{semantically} to embeddings from another encoder for different modality by the mapping. We have carried out an experimental validation of our approach for cross-modal tasks of text-to-image and image-to-text retrieval. Given a text query, the text-to-image mapping translates a text embedding onto the subspace for image embeddings where the translated text embedding can be directly compared to those of images, and vice versa for the image-to-text mapping.  

We can improve the performance of our simple mappings by the choice of pretrained unimodal encoders used. There are off-the-shelf pretrained language models properly biased by contrastive learning with the Natural Language Inference (NLI) dataset. Usually, pretrained image models are already biased properly from training for object classification. An external component such as outer neural layers can further enhance the cross-modal performance. We demonstrate the improved performance by adding an outer multilayer perceptron with gating (gMLP)~\citep{liu2021pay}, a simple neural architecture. 

Our contributions are as follows: i) encoded unimodal representations from off-the-shelf pretrained models can be aligned by a simple, training-free mapping for cross-modal information retrieval; ii) despite simplicity, our cross-modal mappings perform reasonably well reaching the highest accuracy of 77\% on recall@10 comparable to deep neural nets with costly training and fine-tuning; iii) optionally, proper biases introduced by contrastive learning and outer neural architecture such as gated MLP can improve the cross-modal retrieval performance of the proposed mappings. 

%% file: related.tex
\section{Related Work}
Pretraining of deep learning models on large-scale data has flourished under unimodal assumptions. Recently, a self-supervised method makes automated training with unlimited data available on the Internet possible. In computer vision, the success of VGG~\citep{vgg} and ResNet~\citep{he_resnet_2016} is immensely followed while BERT~\citep{bert} and GPT~\citep{brown2020language} have achieved a similar success in NLP. Information retrieval can tremendously benefit from pretrained models although they are valid in a unimodal scenario only.

For the case of cross-modal retrieval, pairing up semantically equivalent data modalities can be considered. To learn a shared embedding subspace, one can explore the idea of jointly training semantically related data with different modalities~\citep{qi_imagebert_2020, huang_pixel-bert_2020, su_vl-bert_2020, li_oscar_2020, sariyildiz_learning_2020, zhang_contrastive_2020,lu_vilbert_2019,tan_lxmert_2019}. Not necessarily for cross-modal information retrieval, these approaches have set language-vision  benchmark tasks and achieved good downstream performances.

Attention mechanism used in masked language modeling can also be applied to a joint image-text encoder as in VL-BERT~\citep{su_vl-bert_2020}. The recent explorations of learning image representations directly from semantic counterparts in natural language have partly inspired us. CLIP~\citep{radford_learning_2021} and ALIGN~\citep{jia_scaling_2021} train images with relevant natural language captions to obtain rich vision-language representations for improving performance on diverse downstream tasks such as text-to-image matching and retrieval. They have demonstrated a simple contrastive learning setup capture better representations without heavily relying on labeled data or a sophisticated neural architecture. To achieve a good performance, however, substantive training effort is inevitable (in a scale of hundreds of high-spec GPUs and billions of training examples).

Instead of laborious and expensive training spent by recent related approaches, we have decided to experiment with simple mappings computable from the least squares and linear projections. The mappings translate a representation from pretrained models in one modality to another such that representations of different modalities can be directly compared for cross-modal retrieval. We take off-the-shelf BERT, RoBERTa, and ViT~\citep{vit} for encoding text and images. The Transformer encoder~\citep{transformer} has originated for NLP, but attention mechanisms have grown their benefits in vision. Transformers are computationally more efficient than convolutional neural nets while providing an on-par or better results for vision tasks. ViT uses the pretrained weights with the JFT-300M dataset~\citep{jft-300m}. Because there is no inductive bias inherent in CNNs such as translation equivariance and locality, the CNN pretraining requires an enormous amount of data if trained from scratch.

%% file: approach.tex
\section{Approach}
\begin{figure}[h!]
\centering
\includegraphics[width=.4\textwidth]{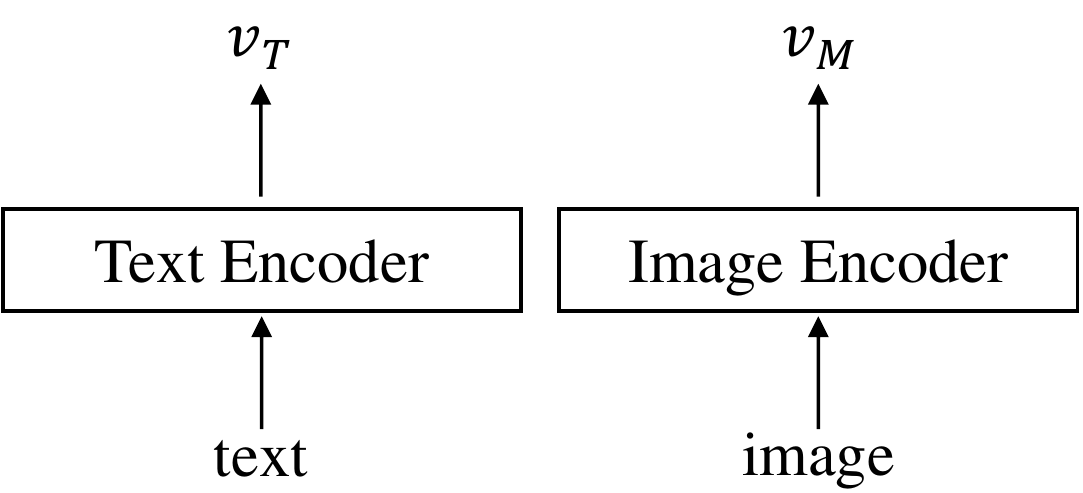}
\caption{Encoders for extracting vectors from text and image.}
\label{fig:encoders}
\end{figure}

Pretrained unimodal encoders take in text and image modalities of data as depicted in Figure~\ref{fig:encoders}. Our goal is to embed one input modality and translate it onto an embedding subspace of the other through a cross-modal mapping. After the translation, embeddings are in the same semantic space and directly compared. That is, the similarity between data examples in different modalities can be examined by computing the inner product of their embedding vectors. In this section, we describe our approach and explain how to compute simple cross-modal mappings. 

\subsection{System of Least Squares via Normal Equations}
Our first method is to learn the text-to-image mapping directly from paired image-text data (and vice versa for image-to-text mapping) via the least squares. Suppose text $T$ and image $M$ that are the source and the target of the mapping (or a linear projection) $\mathbf{\Phi}$. We seek the solution to the problem $\mathbf{V}_T \mathbf{\Phi} = \mathbf{V}_M$ with \begin{align}\nonumber
\mathbf{V}_{T} = \begin{bmatrix}
~\horzbar ~~\mathbf{v}_{T}^{(1)} ~~\horzbar~ \\
~\horzbar ~~\mathbf{v}_{T}^{(2)} ~~\horzbar~ \\
\vdots \\
~\horzbar ~~\mathbf{v}_{T}^{(n)} ~~\horzbar~
\end{bmatrix}
,~
\mathbf{V}_{M} = \begin{bmatrix}
~\horzbar ~~\mathbf{v}_{M}^{(1)} ~~\horzbar~ \\
~\horzbar ~~\mathbf{v}_{M}^{(2)} ~~\horzbar~ \\
\vdots \\
~\horzbar ~~\mathbf{v}_{M}^{(n)} ~~\horzbar~
\end{bmatrix}
,~\\
\mathbf{v}_{T}^{(i)} =\begin{bmatrix}
t_1^{(i)}\\
t_2^{(i)}\\
\vdots \\
t_d^{(i)}
\end{bmatrix}^\top
,~ 
\mathbf{v}_{M}^{(i)} =\begin{bmatrix}
m_1^{(i)}\\
m_2^{(i)}\\
\vdots \\
m_d^{(i)}
\end{bmatrix}^\top
\end{align} where $\mathbf{V}_T$ and $\mathbf{V}_M$ are datasets that contain $n$ embeddings for text $T$ and image $M$ with each $\mathbf{v} \in \mathbb{R}^d$. With $\mathbf{\Phi} = \begin{bmatrix}
\boldsymbol{\phi}^{(1)} & \boldsymbol{\phi}^{(2)}  & \dots & \boldsymbol{\phi}^{(j)} & \dots & \boldsymbol{\phi}^{(d)} 
\end{bmatrix}$ whose element $\boldsymbol{\phi}^{(j)} \in \mathbb{R}^d$ is a column vector, each $\mathbf{V}_T\,\boldsymbol{\phi}^{(j)} = [m_j^{(1)} m_j^{(2)} \dots m_j^{(n)}]$ gives a problem of the least squares. Since $k = 1, \dots, d$, we have a system of $d$ least-square problems that can be solved linear algebraically via the normal equation: $\mathbf{\Phi}^* = \left(\mathbf{V}_{T}^{\top}\mathbf{V}_{T} \right)^{-1}\mathbf{V}_{T}^{\top}\mathbf{V}_{M}$.

\subsection{Solution to the Procrustes Problem} 
Given two data matrices, a source $\mathbf{V}_T$ and a target $\mathbf{V}_M$, the orthogonal Procrustes problem~\citep{procrustes} describes approximation of a matrix searching for an orthogonal projection that most closely maps $\mathbf{V}_T$ to $\mathbf{V}_M$. Formally, we write \begin{align}\label{eq:procrustes}
\mathbf{\Psi}^* = \arg\min_{\mathbf{\Psi}} \left \| \mathbf{V}_T \mathbf{\Psi} - \mathbf{V}_M \right \|_{\textrm{F}}~~~~\textrm{s.t.}~\mathbf{\Psi}^\top\mathbf{\Psi} = \mathbf{I}
\end{align} The solution to Equation~(\ref{eq:procrustes}) has the closed-form $\mathbf{\Psi}^* = \mathbf{X}\mathbf{Y}^\top$ with $\mathbf{X}\mathbf{\Sigma}\mathbf{Y}^\top=\\ \textrm{SVD}(\mathbf{V}_M\mathbf{V}_T^\top)$, where SVD is the singular value decomposition. The Procrustes solution $\mathbf{\Psi}$ gives our second choice for the cross-modal mapping. Similarly, the Procrustes problem for image-to-text mapping can be set up and solved by SVD.  

\subsection{Optional Considerations}
In this section, we describe optional considerations, perhaps with little training, that can improve the baseline cross-modal retrieval performance. Contrastive learning can be set up to minimize the distance between a pair of image and text examples of semantic equivalence (\ie, the text description matches the image)~\citep{zhang_contrastive_2020, radford_learning_2021, jia_scaling_2021, li_oscar_2020}. Conversely, contrastive learning maximizes the distance of a non-matching pair. Because forming non-matching pairs (negatives) can be automated, contrastive learning is convenient to learn the joint representation. Also, it has demonstrated promising results on downstream vision-language tasks. Our approach is differentiated from others over the nonlinear architecture that follows the front-end bimodal (image-text) encoders. We explain our architectural components as follows.

\textbf{gMLP blocks.} We train multi-layer perceptron (MLP) with gating (gMLP)~\citep{liu2021pay}, a nonlinear projection layer following the bimodal encoders. Stacked MLP layers without self-attention are used to capture semantic relationship of image and text data despite having a much lightened network. More importantly, gMLP has exactly the same input and output shape as BERT's and ViT's. This makes sense for using the feature vectors extracted by the bimodal encoders without modification. Our contrastive learning approach emphasizes the training on top of the bimodal encoder output.
 

\textbf{Contrastive learning objective.} From a pair of image and text examples, the bimodal encoders compute $v^{(i)}_T$ and $v^{(i)}_M$ as output to form $\mathcal{D} = \{(v^{(i)}_T, v^{(i)}_M) \}_{i=1}^m$ that are applied to the gMLP nonlinear projection layers. We take the cross-entropy loss for contrastive learning with $N$ in-batch negatives:
\begin{equation}
    \label{eq:objective}
    \begin{aligned}
        \ell_i = -\log \frac{e^{\mathrm{sim}(h_i, h^+_i)/\tau}}{\sum_{j=1}^N e^{\mathrm{sim}(h_i, h_j^+)/\tau}},
    \end{aligned}
\end{equation} where $\tau$ is a temperature hyperparameter, $\mathrm{sim}(h_1,h_2)$ is the cosine similarity $\frac{h_1^\top h_2}{\Vert h_1\Vert \cdot \Vert h_2\Vert}$, and $h_i$ and $h_i^+$ denote output vector of $v^{(i)}_T$ and $v^{(i)}_M$. Optionally, negative pairs can be automatically generated, and the loss function can trivially be modified.

%% file: exp.tex
\section{Experiments}

\begin{table*}[ht]
\centering
\caption{Cross-modal retrieval results on Flickr30K. For comparison, we add the results from CLIP~\citep{radford_learning_2021}. (`+' means an enhanced encoder by biasing.)}
\label{tab:results_tab}
\resizebox{\textwidth}{!}{
\begin{tabular}{c|l|ccccc|ccccc}
\hline
\multirow{2}{*}{Training}                                                                             & \multicolumn{1}{c|}{\multirow{2}{*}{Encoder}} & \multicolumn{5}{c|}{Image to Text}                                            & \multicolumn{5}{c}{Text to Image}                                             \\
                                                                                                      & \multicolumn{1}{c|}{}                       & R@1           & R@5           & R@10          & R@20          & R@100         & R@1           & R@5           & R@10          & R@20          & R@100         \\ \hline
\multirow{4}{*}{\begin{tabular}[c]{@{}c@{}}No\\ Training\\ /GPU\end{tabular}}                         & BERT                                        & 11.5          & 37.6          & 50.5          & 63.3          & 86.8          & 15.5          & 39.4          & 53.4          & 67.0          & 89.4          \\
                                                                                                      & RoBERTa                                     & 18.1          & 44.4          & 58.6          & 70.1          & 90.1          & 17.1          & 42.5          & 56.3          & 70.3          & 91.5          \\
                                                                                                      & BERT+                                       & 29.4          & \textbf{65.2} & 75.9          & 85.5          & \textbf{95.7} & 20.5          & 47.5          & 62.0          & 74.1          & \textbf{93.4} \\
                                                                                                      & RoBERTa+                                    & \textbf{31.9} & \textbf{65.2} & \textbf{77.2} & \textbf{85.8} & 95.0          & \textbf{22.6} & \textbf{51.5} & \textbf{64.4} & \textbf{75.5} & 92.9          \\ \hline
\multirow{4}{*}{\begin{tabular}[c]{@{}c@{}}Outer Layer\\ Training\\ (1 1080ti\\ * 0.3h)\end{tabular}} & BERT                                        & 20.4          & 51.5          & 67.0          & 79.7          & 95.4          & 24.9          & 53.2          & 68.2          & 79.1          & 94.5          \\
                                                                                                      & RoBERTa                                     & 16.3          & 43.1          & 56.5          & 71.4          & 93.4          & 16.4          & 41.4          & 55.6          & 66.7          & 90.4          \\
                                                                                                      & BERT+                                       & \textbf{37.5} & \textbf{71.6} & \textbf{81.5} & \textbf{87.8} & \textbf{97.0} & \textbf{33.4} & \textbf{65.4} & \textbf{77.6} & \textbf{84.9} & \textbf{96.4} \\
                                                                                                      & RoBERTa+                                    & 14.9          & 44.7          & 58.0          & 72.4          & 93.2          & 16.2          & 42.0          & 54.6          & 66.5          & 91.3          \\ \hline

\begin{tabular}[c]{@{}c@{}}Full Encoder\\ Retraining\\ (256 V100\\ * 12 days)\end{tabular}              & CLIP                                        & 88.0          & 98.7          & 99.4          & -             & -             & 68.7          & 90.6          & 95.2          & -             & -             \\ \hline
\end{tabular}}
\end{table*}

\subsection{Setup}
\textbf{Dataset.} To set up cross-modal information retrieval tasks (in both text-to-image and image-to-text directions), we use Flickr30k~\citep{young_flickr_2014} that contains 31,000 images collected from Flickr, each of which is provided with five descriptive sentences by human annotators. A training set of 29,783 pairs and a test set of 1,000 pairs are used.\\
\textbf{Evaluation.} For evaluating the performance of cross-modal retrieval, we use the test partition from Flickr30k. We use the standard evaluation criteria used in most prior work on image-text retrieval task. We adopt recall@$x=1,5,10,20,100$ as our evaluation metric.\\
\textbf{Pretrained encoders.} We use the large BERT~\citep{bert} and RoBERTa~\citep{roberta} models as our text encoders. Both are representative Transformer-based pretrained language models. We also use a large ViT pretrained on the ImageNet-21k dataset as our image encoder. ViT produces the image embeddings (a hidden dimension of 1,024 and a 32x32 patch size that produces 50 hidden vectors) that can be taken in as input to the gMLP layers without any modification.



\subsection{Results}
\begin{figure*}[h]
\centering
\includegraphics[width=1\textwidth]{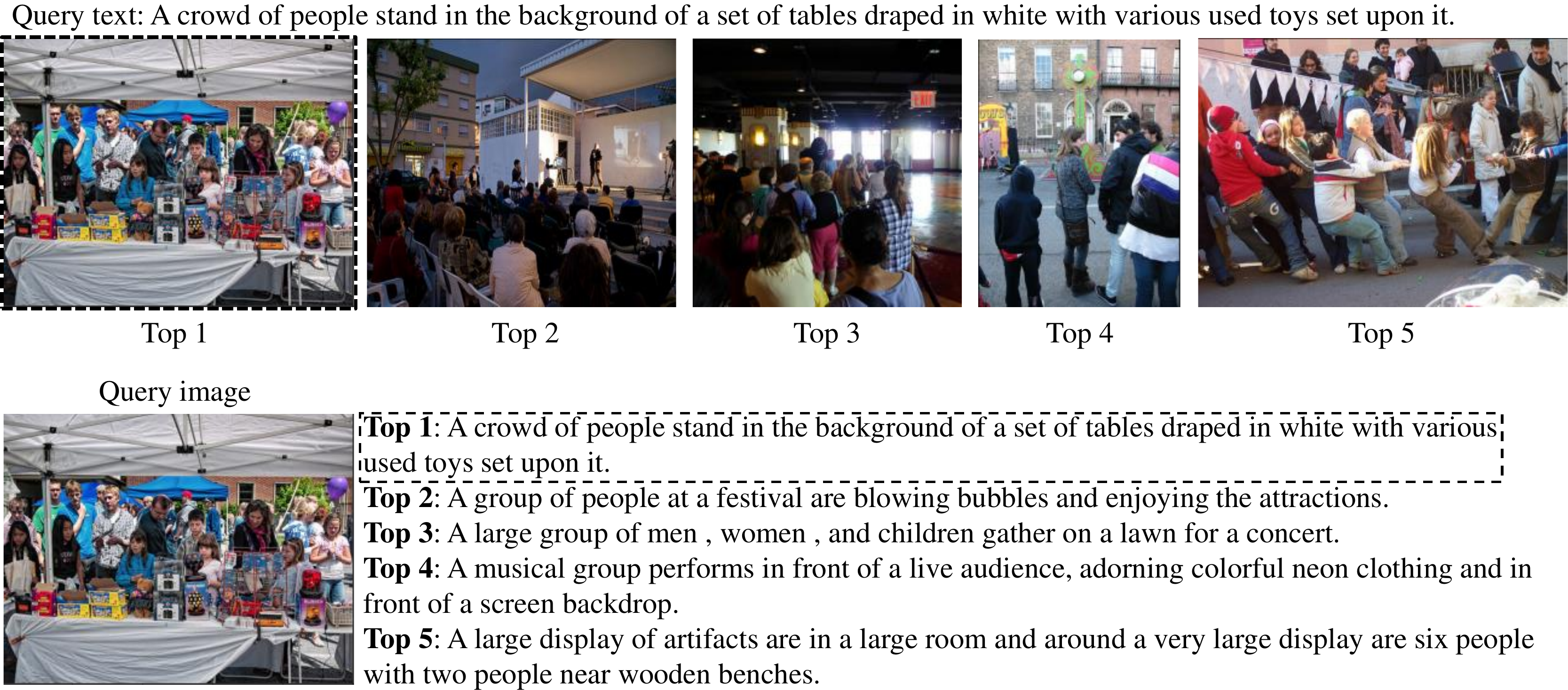}
\caption{Qualitative results of no training model. The dashed line shows the correct retrieval results.}
\label{fig:flick}
\end{figure*}

\textbf{Simple cross-modal mappings.} Our baseline (\i.e., no training) results are presented in Table~\ref{tab:results_tab}. We choose higher number of two different linear mapping methods (least squares and SVD). This simple method reaches 58.6/56.3\% recall@10. We find one of two linear mapping methods fail in some cases. We hypothesize this is caused by poor sentence embeddings of text encoders.\\
\textbf{Using enhanced text encoders via biasing.} The image encoder is originally pretrained to classify 1,000 object classes. On the contrary, the text encoder has not been trained to capture fine-grained sentence representations. \citet{eval_se} show sentence embeddings of BERT without additional learning give poor performance. We hypothesize that this will adversely affect the performance of cross-modal representation. To alleviate the lack of properly contextualized sentence embeddings, we adopt enhanced text encoder (noted BERT+ and RoBERTa+ in Table~\ref{tab:results_tab}) from SimCSE~\citep{gao2021simcse} which uses self-supervised contrastive learning for text encoders. Using enhanced text encoder results in 77.2/64.4\% recall@10 as reported in Table~\ref{tab:results_tab}. Our qualitative results (as shown in Figure~\ref{fig:flick}) show cross-modal retrieval is possible without training.\\
\textbf{Optional gMLP outer layer.} Contrastive learning for the gMLP projection yields the best score of 81.5/77.6\% recall@10. The training takes less than 20 minutes on a single NVIDIA 1080Ti graphics card.

%% file: conc.tex
\section{Conclusion}
With a plethora of large-scale pretrained deep learning models, we have posed an intriguing hypothesis for a light-weight, earth-saving approach to cross-modal information retrieval. Unimodal representations computed by a pretrained model form a high-dimensional embedding subspace of its own. Despite the mess, encoded representations from off-the-shelf pretrained models for different modalities of data can be semantically aligned without additional training. We have formulated classical problems to solve for a simple mapping, which is computable without training but by the least squares and SVD. We have experimented with publicly available pretrained models for text-to-image and image-to-text retrieval tasks. Our simple approach seems to have a good potential for improvement, particularly from the future enhancement of pretrained models. Optionally, if we allow little training to properly bias unimodal encoders and add outer gMLP layers, we can significantly improve the performance of cross-modal information retrieval. 